\def\year{2018}\relax
\documentclass[letterpaper]{article}
\usepackage{aaai18}
\usepackage{times}
\usepackage{helvet}
\usepackage{courier}
\usepackage{url}
\usepackage{graphicx}
\usepackage{latexsym}
\usepackage{amsmath}
\usepackage{amsfonts}
\usepackage{mathrsfs}
\usepackage{xspace}
\usepackage{booktabs}
\usepackage{enumerate}
\DeclareMathAlphabet{\mathscrbf}{OMS}{mdugm}{b}{n}
\title{A Knowledge-Grounded Neural Conversation Model\textsuperscript{*}}
\author{Marjan Ghazvininejad\textsuperscript{1{\scriptsize \raisebox{1pt}{$\dagger$}}}~~~~
Chris Brockett\textsuperscript{2}~~~~
Ming-Wei Chang\textsuperscript{2{\scriptsize \raisebox{1pt}{$\ddagger$}}}\\
{\bf \Large 
Bill Dolan\textsuperscript{2}~~~~
Jianfeng Gao\textsuperscript{2}~~~~
Wen-tau Yih\textsuperscript{2{\scriptsize \raisebox{1pt}{\S}}}~~~~
Michel Galley\textsuperscript{2}}\\
\textsuperscript{1}Information Sciences Institute, USC\\
\textsuperscript{2}Microsoft Research\\
{\tt ghazvini@fb.com, mgalley@microsoft.com}}

\begin{document}
\maketitle

\newcommand{\bleu}{{{\sc BLEU}}\xspace}
\newcommand{\sts}{{{\textsc{Seq2Seq}}}\xspace}
\newcommand{\MTask}{{{\textsc{MTask}}}\xspace}
\newcommand{\MTaskF}{{{\textsc{MTask-F}}}\xspace}
\newcommand{\MTaskR}{{{\textsc{MTask-R}}}\xspace}
\newcommand{\MTaskRF}{{{\textsc{MTask-RF}}}\xspace}
\newcommand\convs[1]{{\bf {#1}}}
\newcommand\slot[1]{{\it {#1}}}

\begin{abstract}
Neural network models are capable of generating extremely natural
sounding conversational interactions.  However, these models have been
mostly applied to casual scenarios (e.g., as ``chatbots'') and have
yet to demonstrate they can serve in more useful conversational
applications.  This paper presents a novel, {\it fully data-driven},
and knowledge-grounded neural conversation model aimed at producing
more contentful responses.  We generalize the widely-used
Sequence-to-Sequence (\sts) approach by conditioning responses on both
conversation history and external ``facts'', allowing the model to be
versatile and applicable in an open-domain setting.  Our approach
yields significant improvements over a competitive \sts baseline.
Human judges found that our outputs are significantly more
informative.
\end{abstract}

\section{Introduction} 

\begin{figure}
\centering
\includegraphics[width=8cm]{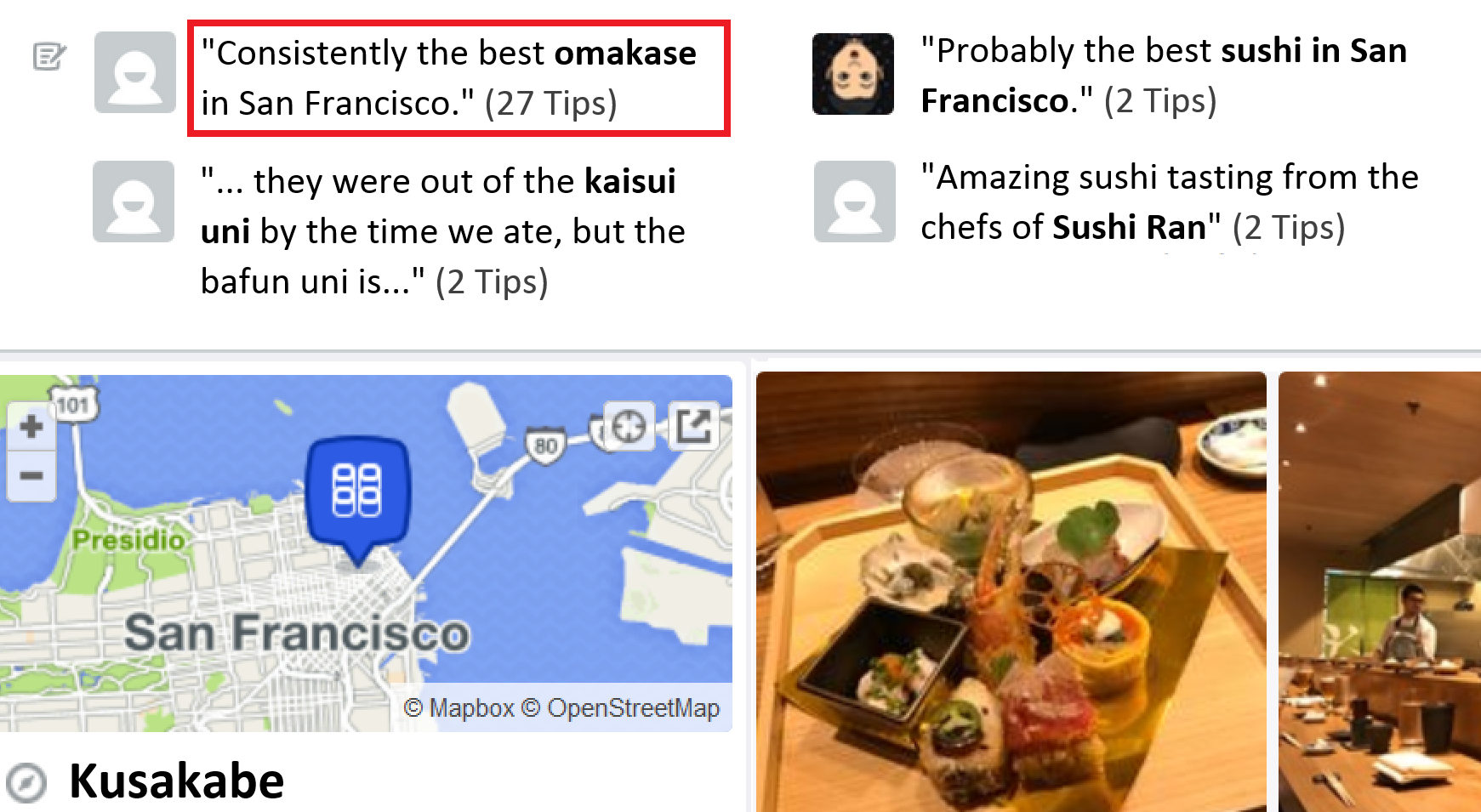}\\[0.2cm]
\begin{small}
\centering
\fbox{\begin{tabular}{l}
{\it \textbf{User input:} Going to Kusakabe tonight.}\\
\textbf{Neural model:} Have a great time! \\

\textbf{Human:} You'll love it! Try omasake, the best in town.~
\end{tabular}}
\end{small}
\caption{Responses of fully data-driven conversation models are often appropriate, but generally lack content characteristic of human responses.}
\label{fig:comp_sent}
\end{figure}

Recent work has shown that conversational chatbot models can be trained in an end-to-end and completely data-driven fashion, without 
hand-coding \cite[{\it inter alia}]{ritter2011data,sordoni2015,shang2015neural,vinyals2015neural,serban2015hierarchical}.
However, fully data-driven systems still lack grounding in the real world and do not have access to external knowledge (textual or structured), which makes it challenging for such systems to respond substantively.
Fig.~\ref{fig:comp_sent} illustrates the difficulty: while an ideal response would  directly reflect on the entities mentioned in the query (user input), neural models produce responses that, while conversationally appropriate,
seldom include factual content.
This contrasts with traditional dialog systems, which can readily inject entities and facts into responses, but often at the cost of significant hand-coding. 
Slot-filler dialog systems are hard put to come up with a natural sounding utterance like the second response in Fig.~\ref{fig:comp_sent} in a manner that is generalizable and scalable. 

\begin{figure*}
\begin{small}
\begin{center}
\begin{tabular}{l}\toprule
A: \convs{Looking forward to trying} \slot{@pizzalibretto} \convs{tonight! my expectations are high}.\\
B: \convs{Get} \slot{the rocco salad}. \convs{Can you eat} \slot{calamari}?\\ \midrule
A: \convs{Anyone in} \slot{Chi} \convs{have a} \slot{dentist office} \convs{they recommend? I'm never going back to} \slot{[...]} \convs{and would love a reco!}\\
B: \convs{Really looved} \slot{Ora} \convs{in} \slot{Wicker Park}.\\\midrule
A: \convs{I'm at} \slot{California Academy of Sciences}\\
B: \convs{Make sure you catch} \slot{the show} \convs{at} \slot{the Planetarium}. \convs{Tickets are usually limited.}\\ \midrule
A: \convs{I'm at} \slot{New Wave Cafe}.\\
B: \convs{Try to get to} \slot{Dmitri's} \convs{for dinner}. \convs{Their} \slot{pan fried scallops} \convs{and} \slot{shrimp scampi} \convs{are to die for.} \\ \midrule
A: \convs{I just bought:} \slot{[...] 4.3-inch portable GPS navigator} \convs{for my wife, shh, don't tell her}.\\
B: \convs{I heard this brand} \slot{loses battery power}. \\ \bottomrule
\end{tabular}
\end{center}
\end{small}
\caption{Social media datasets include many contentful and useful exchanges, e.g., here recommendation dialog excerpts extracted from real tweets. 
While previous models (e.g., \sts{}) succeed in learning the {\bf backbone of conversations}, they have difficulty modeling and producing {\it contentful words} such as named entities, which are sparsely represented in conversation data. To help solve this issue, we rely on non-conversational texts, which represent such entities much more %
exhaustively.}
\label{fig:convo_shape}
\end{figure*}

The goal of this work is to benefit from the versatility and scalability of fully data-driven models, while simultaneously seeking to produce models that are usefully grounded in external knowledge, permitting them to be deployed in, for example, recommendation systems (e.g., for restaurants), and to adapt quickly and easily to new domains.  
The objective here is not task completion as in traditional dialog systems, but the ability to engage a user in a relevant and informative conversation. 
The tie to external data is critical, as the requisite knowledge is often not stored in conversational corpora. 
Much of this information is not found in structured databases either, but is textual, and can be mined from online resources such as Wikipedia, book reviews on Goodreads, and restaurant reviews on Foursquare. 

This paper presents a novel, {\it fully data-driven}, knowledge-grounded neural conversation model aimed at producing contentful responses. Our framework generalizes the Sequence-to-Sequence (\sts{}) approach \cite{hochreiter1997long,sutskever2014sequence} of previous neural conversation models, as it naturally combines conversational and non-conversational data via techniques such as multi-task learning \cite{Caruana:1997,liu-EtAl:2015}.
The key idea is that we can condition responses not only based on conversation history \cite{sordoni2015}, but also on external ``facts'' that are relevant to the current context (for example, Foursquare entries as in Fig.~\ref{fig:comp_sent}). 
Our approach only requires a way to infuse external information based on conversation context (e.g., via simple entity name matching), 
which makes it highly versatile and applicable in an open-domain setting.
Using this framework, we have trained systems at a large scale using 23M general-domain conversations from Twitter and 1.1M Foursquare tips, showing significant improvements in terms of informativeness (human evaluation) over a competitive large-scale \sts{} model baseline. To the best of our knowledge, this is the first large-scale, fully data-driven neural conversation model that effectively exploits external knowledge.

\section{Related Work}

The present work situates itself within the data-driven paradigm of conversation generation, in which statistical and neural machine translation models are derived from conversational data  \cite{ritter2011data,sordoni2015,serban2015hierarchical,shang2015neural,vinyals2015neural,li2016diversity}.
The introduction of contextual models by \cite{sordoni2015} was an important advance within this framework, and we extend their basic approach by injecting side information from textual data.  
Introduction of side information has been shown to be beneficial to machine translation \cite{hoang2016incorporating}, as has also the incorporation of images into multi-modal translation \cite{huang2016attention,delbrouck2017visually}. 
Similarly, \cite{HeHeP17-1162} employ a knowledge graph to embed side information into dialog systems. 
Multi-task learning can be helpful in tasks ranging from query classification to machine translation \cite{Caruana:1997,dong2015multi,liu-EtAl:2015,luong2015multi}. 
We adopt this approach in order to implicitly encode relevant external knowledge from textual data.

This work should be seen as distinct from more goal-directed neural dialog modeling in which question-answer slots are explicitly learned from small amounts of crowd-sourced data, customer support logs, or user data
\cite{wen-EtAl2015,DBLP:journals/corr/WenGMRSVY16,Wen2017Latent,Zhao2017ACL}. 
In many respects, that paradigm
can be characterized as the neural extension of conventional dialog models with or without statistical modeling, e.g., \cite{OhRudnicky2000,Ratnaparkhi2002,BanchsLi2012,AmeixaCoheurEtAl2014}. 
Our purpose is to explore the space of less clearly goal-directed, but nonetheless informative (i.e., informational) conversation that does not demand explicit slot-filling. 

Also relevant is \cite{BordesW16}, who employ memory networks to handle restaurant reservations, using a small number of keywords to handle entity types in a structured knowledge base. 
Similarly \cite{DBLP:conf/eacl/LiuP17a} use memory networks to manage dialog state. 

These works utilize datasets that are relatively small, and unlikely to scale, whereas 
we leverage free-form text to draw on datasets that are several orders of magnitude larger, 
allowing us to cover a greater diversity of domains and forms and thereby learn a more robust conversational backbone.

\section{Grounded Response Generation}

A primary challenge in building fully data-driven conversation models is that most of the world's knowledge is not represented in any existing conversational datasets.
While these datasets \cite{corpora:2015} have grown dramatically in size thanks in particular to social media \cite{ritter2011data}, such datasets are still very far from containing discussions of every entry in Wikipedia, Foursquare, Goodreads, or IMDB. %
This problem considerably limits the appeal of existing data-driven conversation models, as they are bound to respond evasively or deflectively
as in Fig.~\ref{fig:comp_sent}, especially with regard to those entities that are poorly represented in the conversational training data.
On the other hand, even where conversational data representing most entities of interest may exist, we would still face challenges as such huge dataset would be difficult to apply in model training, and many conversational patterns exhibited in the data (e.g., for similar entities) would be redundant.

Our approach aims to avoid redundancy and attempts to better generalize from existing conversational data, as illustrated in Fig.~\ref{fig:convo_shape}.
While the conversations in the figure are about specific venues, products, and services, conversational patterns are general and equally applicable to other entities.
The learned conversational behaviors could be used to, e.g., recommend other products and services. 
A traditional dialog system would use predefined slots to fill conversational backbone (bold text) with content; here, we present a more robust and scalable approach.

\begin{figure}
\centering
\includegraphics[width=8.3cm]{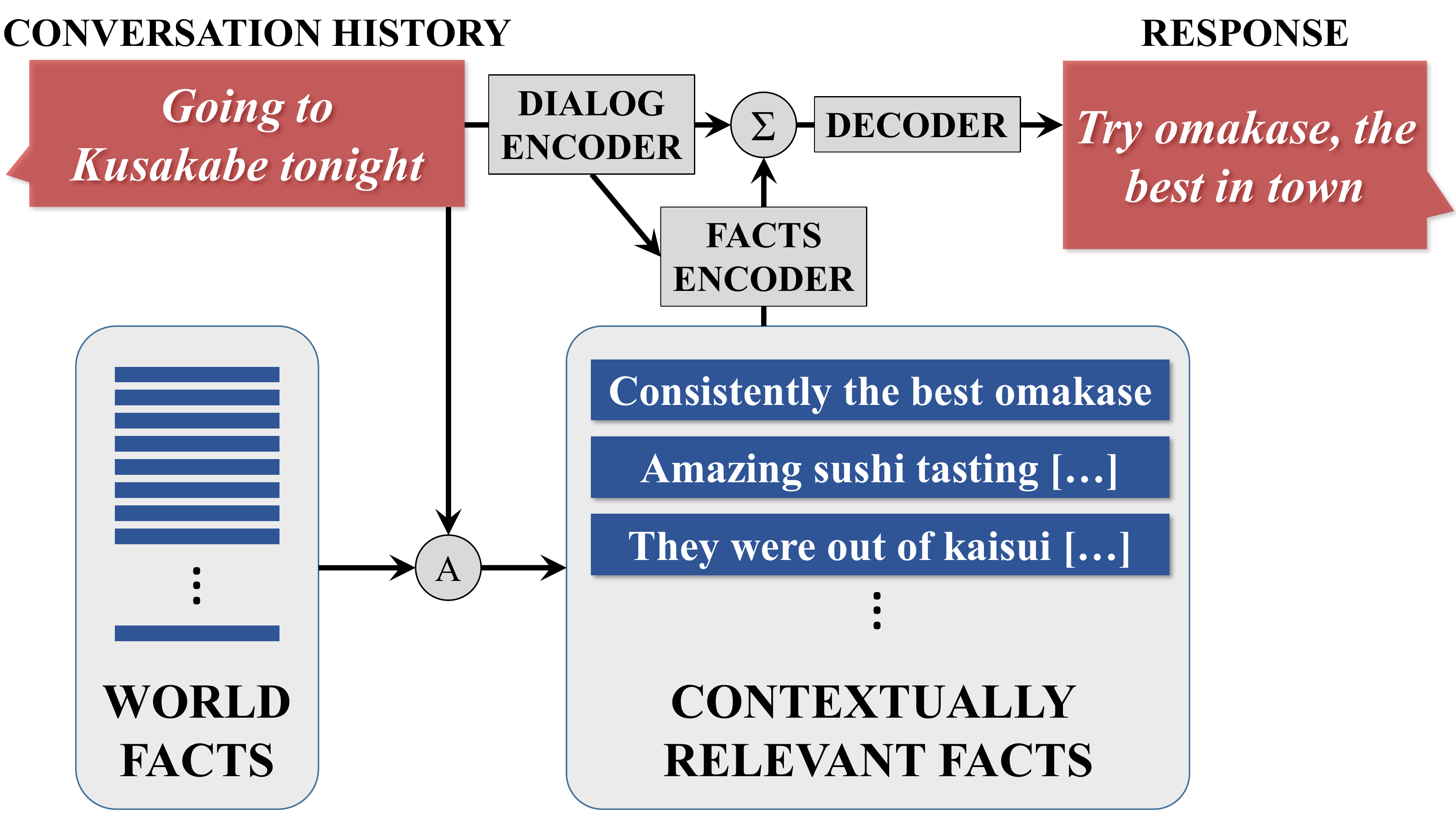}
\caption{Knowledge-grounded model architecture.}
\label{fig:arch}
\end{figure}

In order to infuse the response with factual information relevant to the conversational context, we propose the knowledge-grounded model architecture
depicted in Fig.~\ref{fig:arch}. 
First, we have available a large collection of world facts,\footnote{For presentation purposes, we refer to these items as ``facts'', but a ``fact'' here is simply any snippet of authored text, which may contain subjective or inaccurate information.} which is a large collection of raw text entries (e.g., Foursquare, Wikipedia, or Amazon reviews) indexed by named entities as keys.
Then, given a conversational history or source sequence $S$, we identify the ``focus'' in $S$, which is the text span (one or more entities) based on which we form a query to link to the facts.
This focus can either be identified using keyword matching (e.g., a venue, city, or product name), or detected using more advanced methods such as entity linking or named entity recognition. 
The query
is then used to retrieve all contextually relevant facts: \mbox{$F=\{f_1,...,f_{k}\}$}.\footnote{In our work, we use a simple keyword-based IR engine to retrieve relevant facts from the full collection (see Datasets section).} 
Finally, both conversation history and relevant facts are fed into a neural architecture
that features distinct encoders for conversation history and facts. We will detail this architecture in the subsections below.

This knowledge-grounded approach is more general than \sts response generation, as it avoids the need to learn the same conversational pattern for each distinct entity that we care about. 
In fact, even if a given entity (e.g., @pizzalibretto in Fig.~\ref{fig:convo_shape}) 
is not part of our conversational training data and is therefore out-of-vocabulary, our approach is still able to rely on retrieved facts to generate an appropriate response.
This also implies that we can enrich our system with new facts without the need to retrain the full system.

We train our system using multi-task learning \cite{Caruana:1997,luong2015multi} as a way of combining conversational data that is naturally associated with external data (e.g., discussions about restaurants and other businesses as in Fig.~\ref{fig:convo_shape}), and less
informal
exchanges (e.g., a response to {\it hi, how are you}).
More specifically, our multi-task setup contains two types of tasks:
\begin{enumerate}[{(1)}]
\item one purely conversational, where we expose the model without fact encoder to $(S,R)$ training examples, $S$ representing the conversation history and $R$ the response;
\item the other task exposes the full model with $(\{f_1,\ldots,f_k,S\},R)$ training examples.
\end{enumerate}
This decoupling of the two training conditions offers several advantages, including:
First, it allows us to pre-train the conversation-only dataset separately, and start multi-task training (warm start) with a dialog encoder and decoder that already learned the backbone of conversations.
Second, it gives us the flexibility to expose different kinds of conversational data in the two tasks.
Finally, one interesting option is to replace the response in task (2) with one of the facts ($R=f_i$), which makes task (2) similar to an autoencoder and helps produce responses that are even more contentful.

\subsection{Dialog Encoder and Decoder}

The dialog encoder and response decoder form together a sequence-to-sequence (\sts model \cite{hochreiter1997long,sutskever2014sequence}, which has been successfully used in building end-to-end conversational systems \cite{sordoni2015,vinyals2015neural,li2016diversity}. Both encoder and decoder are recurrent neural network (RNN) models: an RNN that encodes a variable-length input string into a fixed-length vector representation and an RNN that decodes the vector representation into a variable-length output string. This part of our model is almost identical to prior conversational \sts models, except that we use gated recurrent units (GRU) \cite{gru:2014} instead of LSTM \cite{hochreiter1997long} cells. Encoders and decoders in  the present model do not share weights or word embeddings. 

\subsection{Facts Encoder}

The Facts Encoder of Fig.~\ref{fig:arch} is similar to the Memory Network model first proposed by \cite{weston2014,sukhbaatar2015}.
It uses an associative memory for modeling the facts 
relevant to a particular problem---in our case, an entity mentioned in a conversation--then retrieves and weights these facts based on the user input and conversation history to generate an answer. 
Memory network models 
have been successfully used 
in Question Answering to make inferences based on the facts saved in the memory \cite{weston2015}.

In our adaptation of memory networks, we use an RNN encoder to turn the input sequence (conversation history) into a vector, instead of a bag of words representation as used in the original memory network models.  
This enables us to better exploit interlexical dependencies between different parts of the input, and makes this memory network model (facts encoder) more directly comparable to a \sts model.

More formally, we are given an input sentence $S=\{s_{1}, s_{2}, ..., s_{n}\}$, and a fact set $F=\{f_1,f_2,...,f_k\}$ that are relevant to the conversation history. 
The RNN encoder reads the input string word by word and updates its hidden state. 
After reading the whole input sentence the hidden state of the RNN encoder, $u$ is the summary of the input sentence. 
By using an RNN encoder, we have a rich representation for a source sentence.

Let us assume $u$ is a $d$ dimensional vector and $r_i$ is the bag of words representation of $f_i$ with dimension $v$. Based on \cite{sukhbaatar2015} we have:
\begin{align}
m_i=A r_i\\
c_i=C r_i\\
p_i=\textrm{softmax}(u^T m_i)\\
o=\sum _{i=1}^{k} p_i c_i\\
\hat u = o+u \label{eq:sum}
\end{align}\\[-0.5cm]
Where $A,C \in \mathbb{R}^{d\times v}$ are the parameters of the memory network. Then, unlike the original version of the memory network, we use an RNN decoder that is good for generating the response. The hidden state of the RNN is initialized with $\hat u$ which is a symmetrization of input sentence and the external facts, to predict the response sentence $R$ word by word.

As alternatives to summing up facts and dialog encodings in equation~\ref{eq:sum}, we also experimented with other operations such as concatenation, but summation seemed to yield the best results.
The memory network model of \cite{weston2014} can be defined as a multi-layer structure. 
In this task, however, 1-layer memory network was used, since multi-hop induction was not needed.

\section{Datasets}

The approach we describe above is quite general, and is applicable to any dataset that allows us to map named
entities to free-form text (e.g., Wikipedia, IMDB, TripAdvisor, etc.).
For experimental purposes, we utilize datasets derived from two popular social media services: 
Twitter
 (conversational data) and 
Foursquare
 (non-conversational data).

\paragraph{Foursquare:}
Foursquare tips are comments left by customers about restaurants and other, usually commercial, establishments. 
A large proportion of these describe aspects of the establishment, and provide recommendations about what the customer enjoyed (or otherwise)
We extracted from the web 1.1M tips relating to establishments in North America. 
This was achieved by identifying a set of 11 likely ``foodie'' cities and then collecting tip data associated with zipcodes near the city centers. 
While we targeted foodie cities, the dataset is very general and contains tips applicable to many types of local businesses (restaurants, theaters, museums, stores, etc.) 
In the interests of manageability for experimental purposes, we ignored establishments associated with fewer than 10 tips, but other experiments with up to 50 tips per venue yield comparable results.
We limited the tips to those for which Twitter handles were found in the Twitter conversation data.

\paragraph{Twitter:}
We collected a {\bf 23M general dataset}
of \mbox{3-turn} conversations.
This serves as a background dataset not associated with facts, and its massive size is key to learning the conversational structure or backbone.

Separately, on the basis of Twitter handles found in the Foursquare tip data, we collected approximately 1~million two-turn conversations that contain entities that tie to Foursquare. We refer to this as
the {\bf 1M grounded dataset}.
Specifically, we identify conversation pairs 
in which the first turn contained either a handle of the business name (preceded by the ``@'' symbol) or a hashtag that matched a handle.\footnote{This mechanism of linking conversations to facts using exact match on the handle is high precision but low recall, but low recall seems reasonable as we are far from exhausting all available Twitter and Foursquare data.}
Because we are interested in conversations among real users (as opposed to customer service agents), we removed conversations where the response was generated by a user with a handle found in the Foursquare data.

\subsection{Grounded Conversation Datasets}

We augment the 1M grounded dataset with facts (here Foursquare tips) relevant to each conversation history. The number of contextually relevant tips for some handles can sometimes be enormous, up to 10k. To filter them for relevance to the input, the system vectorizes the input (as tf-idf weighted word counts) and each of the retrieved facts, and calculates cosine similarity between the input sentence and each of the tips and retains $10$ tips with the highest score.

Furthermore, for a significant portion of the 1M Twitter conversations collected using handles found on Foursquare, the last turn was not particularly informative, e.g., when it provides a purely socializing response (e.g., {\it have fun there}).
As one of our goals is to evaluate conversational systems
on their ability to produce {\it contentful} responses, we select
a dev and test set (4k conversations in total) designed to contain
responses that are informative and useful.

For each handle, we created two scoring functions: 
\begin{itemize}
\item{Perplexity according to a 1-gram LM trained on all the tips containing that handle.}
\item{$\chi$-square score, which measures how much content each token bears in relation to the handle. Each tweet is then scored according to the average content score of its terms.}
\end{itemize}
In this manner, we selected 15k top-ranked conversations using the LM score and 15k using the chi-square score.
A further 15k conversations were randomly sampled. 
We then randomly sampled 10k conversations from these 45K conversations.  
Crowdsourced human judges were then presented with these 10K sampled conversations and asked to determine whether the response contained actionable information, i.e., did they contain information that would permit the respondents to decide, e.g., whether or not they should patronize an establishment. 
From this, we selected the top-ranked 4k conversations to be held out as validation set and test set; these were removed from our training data.   

\section{Experimental Setup}

\subsection{Multi-Task Learning}

We use multi-task learning with these tasks:
\begin{itemize}
    \item \textsc{Facts} task: We expose the full model to $(\{f_1,...,f_n,S\},R)$ training examples.
    \item \textsc{NoFacts} task: We expose the model without fact encoder to $(S,R)$ examples.
    \item \textsc{Autoencoder} task: This is similar to the \textsc{Facts} task, except that we replace the response with each of the facts, i.e., this model is trained on $(\{f_1,...,f_n,S\},f_i)$ examples.
    There are $n$ times many samples for this task than for the \textsc{Facts} task.\footnote{This is akin to an autoencoder as the fact $f_i$ is represented both in the input and output, but is of course not strictly an autoencoder.}
\end{itemize}

The tasks \textsc{Facts} and \textsc{NoFacts} are representative of how our model is intended to work, but we found that the \textsc{Autoencoder} tasks helps inject more factual content into the response. The different variants of our multi-task learned system exploit these tasks as follows:
\begin{itemize}
   \item \sts{}: Trained on task \textsc{NoFacts} with the 23M general conversation dataset. Since there is only one task, it is not {\it per se} a multi-task setting.
   \item \MTask: Trained on two instances of the \textsc{NoFacts} task, respectively with the 23M general dataset and 1M grounded dataset (but without the facts). While not an interesting system in itself, we include it to assess the effect of multi-task learning separately from facts.
    \item \MTaskR: Trained on the \textsc{NoFacts} task with the 23M dataset, and the \textsc{Facts} task with the 1M grounded dataset.
    \item \MTaskF: Trained on the \textsc{NoFacts} task with the 23M dataset, and the \textsc{Autoencoder} task with the 1M dataset.
    \item \MTaskRF: Blends \MTaskF and \MTaskR, as it incorporates 3 tasks:
    \textsc{NoFacts} with the 23M general dataset, 
    \textsc{Facts} with the 1M grounded dataset, 
    and \textsc{Autoencoder} again with the 1M dataset.
\end{itemize}

We trained a one-layer memory network structure with two-layer \sts models.
More specifically, 
we used 2-layer GRU models with 512 hidden cells for each layer for encoder and decoder,
the dimensionality of word embeddings is set to 512, 
and the size of input/output memory representation is 1024.
We used the Adam optimizer with a fixed learning rate of 0.1.
Batch size is set to 128.
All parameters are initialized from a uniform distribution in $[-\sqrt {3/d}, \sqrt {3/d} ]$, where $d$ is the dimension of the parameter.
Gradients are clipped at 5 to avoid gradient explosion.

Encoder and decoder use different sets of parameters. The top 50k frequent types from conversation data is used as vocabulary which is shared between both conversation and non-conversation data.
We use the same learning technique as \cite{luong2015multi} for multi-task learning. In each batch, all training data is sampled from one task only. For task $i$ we define its mixing ratio value of $\alpha_i$, and for each batch we select randomly a new task $i$ with probability of ${\alpha_i}/{\sum_j \alpha_j}$ and train the system by its training data.

\subsection {Decoding and Reranking}

We use a beam-search decoder similar to~\cite{sutskever2014sequence} with beam size of 200, and maximum response length of 30. 
Following \cite{li2016diversity}, we generate $N$-best lists containing three features: (1) the log-likelihood $\log P(R|S,F)$ according to the decoder; (2) word count; (3) the log-likelihood $\log P(S|R)$ of the source given the response.
The third feature is added to deal with the issue of generating commonplace and generic responses such as {\it I don't know}, 
which is discussed in detail in \cite{li2016diversity}.
Our models often do not need the third feature to be effective, but---since our baseline needs it to avoid commonplace responses---we include this feature in all systems.
This yields the following reranking score:
\begin{equation*}
\log P(R|S,F) + \lambda \log P(S|R) + \gamma |R|
\end{equation*}
$\lambda$ and $\gamma$ are free parameters, which we tune on 
our development $N$-best lists
using MERT \cite{mert}
by optimizing \bleu. To estimate $P(S|R)$ we train a Sequence-to-sequence model  by swapping messages and responses. In this model we do not use any facts.

\subsection{Evaluation Metrics}

Following \cite{sordoni2015,li2016diversity,Wen2017Latent}, we use \bleu automatic evaluation. While \cite{notbleu:2016} suggest that \bleu correlates poorly with human judgment at the sentence-level,\footnote{This corroborates earlier findings that accurate sentence-level automatic evaluation is indeed difficult, even for Machine Translation \cite{graham-baldwin-mathur:2015:NAACL-HLT}, as \bleu and related metrics were originally designed as corpus-level metrics.} we use instead corpus-level \bleu, which is known to better correlate with human judgments \cite{MetricsMATR:2008}, including for response generation \cite{galley2015}. We also report perplexity and lexical diversity, the latter as a raw yet automatic measure of informativeness and diversity.
Automatic evaluation is augmented with human judgments of appropriateness and informativeness.

\section{Results}

\begin{table}
\centering
{\small
\begin{tabular}{l|c|c}\toprule
    &       \multicolumn{2}{c}{Perplexity} \\
 \multicolumn{1}{c|}{Model} & \multicolumn{1}{c|}{General Data} & \multicolumn{1}{c}{Grounded Data} \\ %
 \midrule
\sts       & {\bf 55.0} & 214.4 \\
\sts-S      & 125.7 & {\bf 82.6 }\\
\midrule 
\MTask      & 57.2 & 82.5 \\
\MTaskR     & {\bf 55.1}  & {\bf 77.6 }\\
\MTaskF     & 77.3 & 448.8 \\
\MTaskRF    & 67.2 & 97.7  \\
\bottomrule
\end{tabular}}
\caption{Perplexity of different models. \sts-S is a \sts model that is trained on the \textsc{NoFacts} task with 1M grounded dataset (without the facts). }
\label{tab:perplexity} 
\end{table}

\begin{table}
\centering
{\small
\begin{tabular}{l|c|r|r}
\toprule
                & & \multicolumn{2}{c}{Diversity} \\
\multicolumn{1}{c|}{Model}   & \multicolumn{1}{c|}{BLEU} & \multicolumn{1}{c|}{1-gram} & \multicolumn{1}{c}{2-gram} \\
\midrule
\sts               &  0.55 &  4.14\%  &  14.4\% \\
\MTask             &  0.80 &  2.35\%  &   5.9\% \\
\midrule
\MTaskF            &  0.48 &  9.23\%  &  26.6\% \\
\MTaskR            &  {\bf 1.08} &  7.08\%  &  21.9\% \\
\MTaskRF           &  0.58 &  {\bf 8.71}\%  & {\bf 26.0}\% \\
 \bottomrule
\end{tabular}}
\caption{BLEU-4 and lexical diversity.}
\label{tab:bleu} 
\end{table}

\paragraph{Automatic Evaluation:}
We computed perplexity and \bleu \cite{Papineni2002BLEU} for each system. These are shown in Tables \ref{tab:perplexity} and \ref{tab:bleu} respectively. We notice that the \sts model specifically trained on general data has high perplexity on grounded data.\footnote{Training the system on just 1M grounded data with \textsc{Facts} doesn't solve this problem, as its perplexity on general data is also high (not in table).}
We observe that the perplexity of \MTask and \MTaskR models on both general and grounded data is as low as the \sts models that are trained specifically on general and grounded data respectively. As expected, injecting more factual content into the response in  \MTaskF and \MTaskRF increased the perplexity especially on grounded data. 

\bleu scores are low, but this is not untypical of conversational systems (e.g., \cite{li2016diversity,li2016persona}). Table~\ref{tab:bleu} shows that the \MTaskR model yields
a significant performance boost, with a \bleu score increase of $96\%$ and $71\%$ jump in 1-gram diversity compared to the competitive \sts baseline. In terms of \bleu scores, \MTaskRF improvements is not significant, but it generates the highest 1-gram and 2-gram diversity among all models.

\paragraph{Human Evaluation:} We crowdsourced human evaluations. We had annotators judge 500 randomly-interleaved paired conversations, asking them which was better on two parameters: appropriateness to the  context, and informativeness. The crowd workers were instructed to: \textit{Decide which response is more appropriate, i.e., which is the best conversational fit with what was said. Then decide which of the two is more informative (i.e., knowledgeable, helpful, specific) about the establishment under discussion}. 
Judges were asked to select among \textit{Clearly{\#}1}, \textit{Maybe {\#}Number 1}, \textit{About the Same}, \textit{Maybe {\#}2}, and \textit{Clearly {\#}2}. These were converted to scores between 1 and 0, and assigned to the pair members depending on the order in which the pair was presented.
Seven judges were assigned to each pair.\footnote{Annotators whose variance fell greater than two standard deviations from the mean variance were dropped.}

\begin{table*}
\centering
{\small
\begin{tabular}{lrrrrrrrr}
\toprule
Comparison 
                  & \multicolumn{4}{ |c }{Informativeness}  
                  & \multicolumn{4}{ |c }{Appropriateness} 
                  \\[1pt]
\midrule
\sts vs \MTask    
                  & \multicolumn{1}{ |r }{ 0.501}      & $\pm$0.016 &  0.499 &  $\pm$0.015 
                  & \multicolumn{1}{ |r }{\bf{0.530}}  &  \bf{$\pm$0.017} &  0.470 &  $\pm$0.017 
                  \\ [1pt]
\sts  vs \MTaskF  
                  & \multicolumn{1}{ |r }{0.478}       & $\pm$0.015 &  \bf{0.522} & \bf{$\pm$0.015}  
                  & \multicolumn{1}{ |r }{\bf{0.537}}  & \bf{$\pm$0.016} &  0.463 &  $\pm$0.017 
                  \\ [1pt]
\sts  vs \MTaskRF 
                  & \multicolumn{1}{ |r }{ 0.492}      & $\pm$0.013 &  0.508 &  $\pm$0.013 
                  & \multicolumn{1}{ |r }{ 0.502}      &  $\pm$0.015 &  0.498 & $\pm$0.014 
                  \\ [1pt]
\midrule
\sts  vs \MTaskR(*)  
                  & \multicolumn{1}{ |r }{0.479}       & $\pm$0.017 &  \bf{0.521} &  \bf{$\pm$0.013}
                  & \multicolumn{1}{ |r }{0.495}       &  $\pm$0.015 &  0.505 & $\pm$0.015 
                  \\ [1pt]
 \bottomrule
\end{tabular}
}
\caption{Mean differences in judgments in human evaluation, together with 95\% confidence intervals. Differences sum to 1.0. Boldface items are significantly better (p \textless 0.05) than their comparator. (*): Main system, pre-selected on the basis of BLEU.}
\label{tab:UHRS-Eval} 
\end{table*}

The results of annotation are shown in Table~\ref{tab:UHRS-Eval}. 
Our primary system \MTaskR, which performed best on BLEU, significantly outperforms the \sts baseline on \textit{Informativeness} (p = 0.003) and shows a small, but non-statistically-significant gain with respect \textit{Appropriateness}.
Other systems are included in the table for completeness.
The ``vanilla'' \MTask shows no significant gain in \textit{Informativeness}.
\MTaskF performed significantly better than baseline (p = 0.005) on \textit{Informativeness}, but was significantly worse on \textit{Appropriateness}. 
\MTaskRF came in slightly better than baseline on \textit{Informativeness} but worse on \textit{Appropriateness}, though in neither case is the difference statistically significant by the conventional standard of $\alpha = 0.05$. 
In sum, our best performing \MTaskR system appears to have successfully balanced the needs of informativeness and maintaining contextual appropriateness.  

\begin{figure}
\centering 
\includegraphics[width=0.42\textwidth]{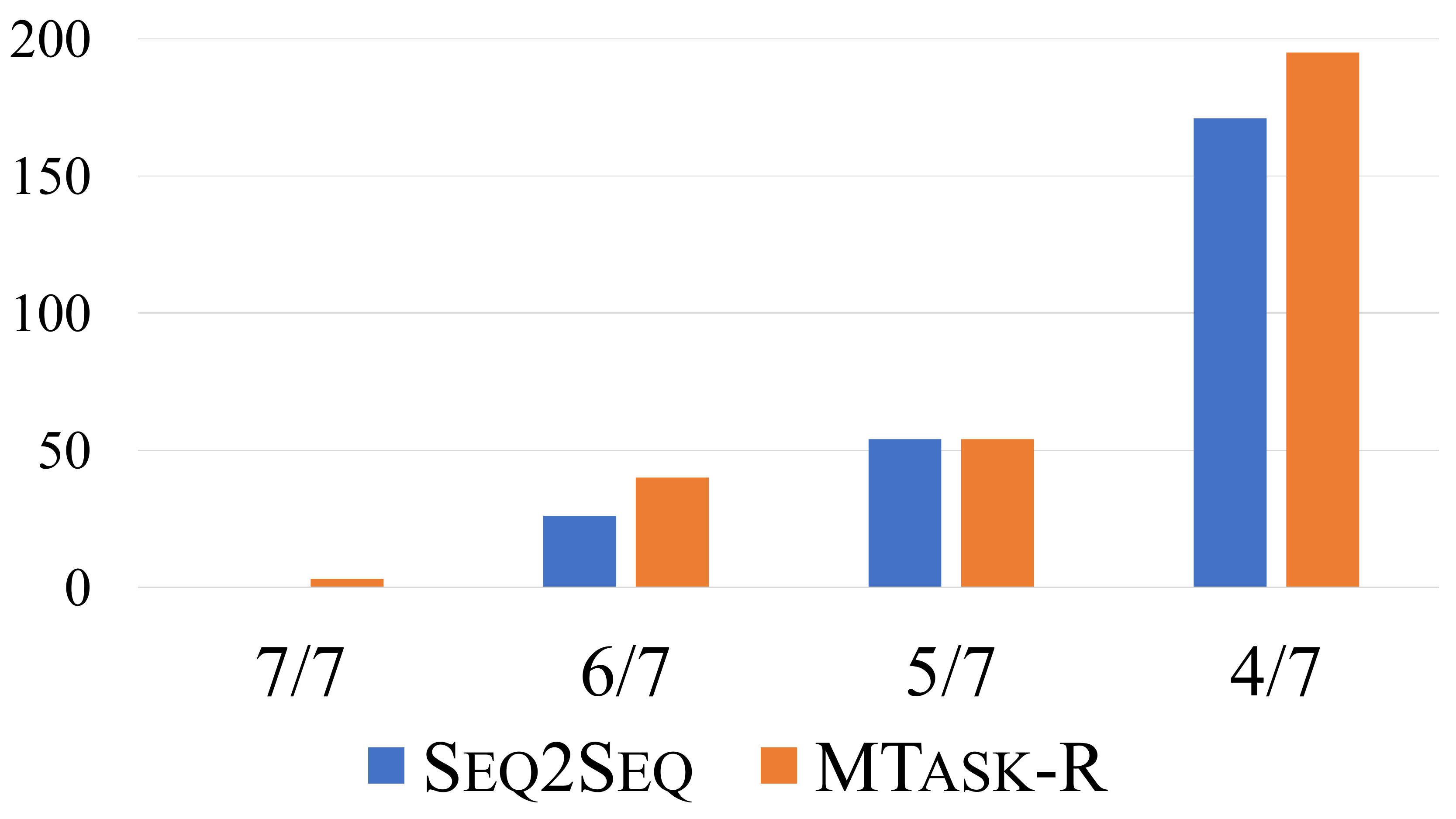}
\caption{Judge preference counts (appropriateness) for \MTaskR versus \sts.} 
\label{fig:mem-tweets-app}
\end{figure}

\begin{figure}
\centering 
\includegraphics[width=0.42\textwidth]{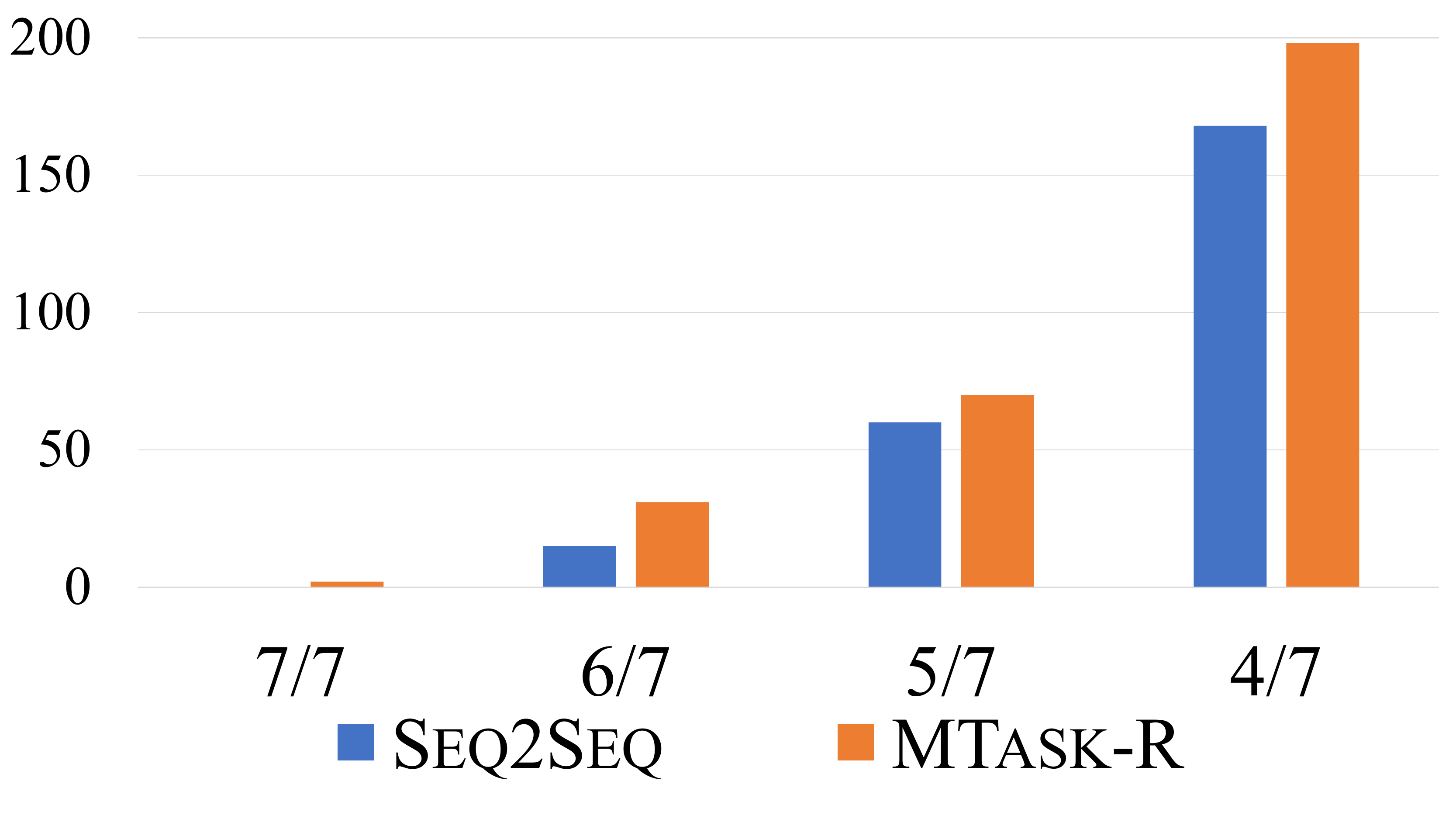}
\caption{Judge preference counts (informativeness) for \MTaskR versus \sts.} 
\label{fig:mem-tweets-inf}
\end{figure}

The narrow differences in averages in Table \ref{tab:UHRS-Eval} tend to obfuscate the judges' voting trends. 
To clarify the picture, we translated the scores into the ratio of judges who preferred that system and binned the counts.  
Figs. \ref{fig:mem-tweets-app} and \ref{fig:mem-tweets-inf} compare \MTaskR with the \sts baseline. 
Bin 7 on the left corresponds to the case where all 7 judges ``voted'' for the system, bin 6 to that where 6 out of 7 judges ``voted'' for the system, and so on.\footnote{Partial scores were rounded up,  affecting both systems equally.} 
Other bins are not shown since these are a mirror image of bins 7 through 4. 
The distributions in Fig. \ref{fig:mem-tweets-inf} are sharper and more distinctive than in  Fig. \ref{fig:mem-tweets-app}. indicating that judge preference for the \MTaskR model is relatively stronger when it comes to informativeness. 
 
\section{Discussion}

\begin{figure*}[ht]
\centering
\begin{small}
\begin{tabular}{ll}\toprule
1& A: I would give ALMOST anything for some [{\it Mexican restaurant}] right now.\\ 
& B: Me too. Creamy jalapeno sauce is really good.\\
\midrule
2& A: Visiting the celebs at Los Angeles International Airport (LAX) - [...] w/ 70 others\\
& B: Nice airport terminal. Have a safe flight.\\
\midrule
3& A: Obsessed with [{\it jewelry company}] :-*\\
& B: oh my gosh obsessed with their bracelets and the meaning behind them !\\
\midrule
4&A: Live right now on [{\it FM radio}] Tune in!!!!!\\
&B: Listen to Lisa Paige\\
\midrule
5& A: Is that [...] in your photos? It's on my list of places to visit in NYC.\\
& B: Don't forget to check out the 5th floor, while you are here, it's a great view.\\
\midrule
6& A: Are you [{\it coworking community}] today or tomorrow?\\
& B: I think so. Try the cookies and coffee.\\
\midrule
7&A: Cant wait for my [{\it pizzeria}] tonight\\
&B: Order bacon n pepperoni thin crust pizza.\\
\midrule
8&A: So [...] is down to one copy of Pound Foolish. I'm curious to see if they are re-ordering it.\\
&B: Check out the video feed on 6 and take a picture of the Simpsons on the 3rd floor.\\
\midrule
\midrule
9&A: I wish [...] would introduce another vegetarian option besides the shroomburger. It's delicious but kind of ridiculous.\\
&B: This is the best j.crew in the world. Try the lemonade!\\
\midrule
10&A: Just had an awesome dinner at [{\it restaurant}] Great recommendation [{\it person}]\\
&B: One of my favorite places I've ever been to in NYC. The food is great and the service is lackluster.\\
\bottomrule
\end{tabular}
\end{small}
\caption{Responses produced by our neural response generation system \MTaskRF. Entities marked [...] are anonymized.}
\label{fig:examples}
\end{figure*}

Figure~\ref{fig:examples} presents examples from the \MTaskRF model, and illustrates that our responses are generally both appropriate and informative. First, we note that our models preserve the ability of earlier work \cite{sordoni2015,vinyals2015neural} to respond contextually and appropriately on a variety of topics, with responses such as {\it me too} (1) and {\it have a safe flight} (2). Second, our grounded models often incorporate information emerging from ``facts'', while usually keeping the responses contextually appropriate. For example in (3), those facts revolve mostly around jewelry such as bracelets, which leads the system to respond {\it obsessed with their bracelets and the meaning behind them}, while {\it meaning behind them} does not belong to any of the facts and is instead ``inferred'' by the model (which associates jewelry with sentimental value). Responses influenced mostly by facts may occasionally contain a single unedited fact (4-5), but otherwise generally combine text from different facts (6-8).\footnote{Facts: {\it grab a cup of coffee and get productive} and {\it try the cookies in the vending machine of local food} (6); {\it sit with and take a picture of the Simpsons on the 3rd floor} and {\it Check out the video feed on 6 and Simpsons/billiards on 3!} (8).} Examples 9 and 10 are negative examples that illustrate the two main causes of system errors: the presence of an irrelevant fact (e.g., {\it j.crew } in example 9), and the system combining self-contradictory facts (10).  Despite such errors, judges found that our best grounded system is generally on par with the \sts{} system in terms of appropriateness, while significantly improving informativeness (Table~\ref{tab:UHRS-Eval}).

\section{Conclusions}

We have presented a novel knowledge-grounded conversation engine that could serve as the core component of a multi-turn recommendation or conversational QA system.
The model is a large-scale, scalable, fully data-driven neural conversation model that effectively exploits external knowledge, and does so without explicit slot filling.
It generalizes the \sts{} approach to neural conversation models by naturally combining conversational and non-conversational data through multi-task learning. 
Our simple entity matching approach to grounding external information based on conversation context makes for a model that is informative, versatile and applicable in open-domain systems. 

\section*{Acknowledgements}
We thank Xuetao Yin and Leon Xu for helping us obtain Foursquare data, and Kevin Knight, Chris Quirk, Nebojsa Jojic, Lucy Vanderwende, Vighnesh Shiv, Yi Luan, John Wieting, Alan Ritter, Donald Brinkman, and Puneet Agrawal.

\bibliography{main}

\begin{thebibliography}{}

\bibitem[\protect\citeauthoryear{Ameixa \bgroup et al\mbox.\egroup
  }{2014}]{AmeixaCoheurEtAl2014}
Ameixa, D.; Coheur, L.; Fialho, P.; and Quaresma, P.
\newblock 2014.
\newblock Luke, {I} am your father: dealing with out-of-domain requests by
  using movies subtitles.
\newblock In {\em Intelligent Virtual Agents}.

\bibitem[\protect\citeauthoryear{Banchs and Li}{2012}]{BanchsLi2012}
Banchs, R.~E., and Li, H.
\newblock 2012.
\newblock {IRIS}: a chat-oriented dialogue system based on the vector space
  model.
\newblock {\em ACL}.

\bibitem[\protect\citeauthoryear{Bordes and Weston}{2017}]{BordesW16}
Bordes, A., and Weston, J.
\newblock 2017.
\newblock Learning end-to-end goal-oriented dialog.
\newblock {\em ICLR 2017}.

\bibitem[\protect\citeauthoryear{Caruana}{1997}]{Caruana:1997}
Caruana, R.
\newblock 1997.
\newblock Multitask learning.
\newblock {\em Machine Learning} 28(1):41--75.

\bibitem[\protect\citeauthoryear{Chung \bgroup et al\mbox.\egroup
  }{2014}]{gru:2014}
Chung, J.; G{\"{u}}l{\c{c}}ehre, {\c{C}}.; Cho, K.; and Bengio, Y.
\newblock 2014.
\newblock Empirical evaluation of gated recurrent neural networks on sequence
  modeling.
\newblock {\em CoRR} abs/1412.3555.

\bibitem[\protect\citeauthoryear{Delbrouck, Dupont, and
  Seddati}{2017}]{delbrouck2017visually}
Delbrouck, J.-B.; Dupont, S.; and Seddati, O.
\newblock 2017.
\newblock Visually grounded word embeddings and richer visual features for
  improving multimodal neural machine translation.
\newblock In {\em Grounding Language Understanding workshop}.

\bibitem[\protect\citeauthoryear{Dong \bgroup et al\mbox.\egroup
  }{2015}]{dong2015multi}
Dong, D.; Wu, H.; He, W.; Yu, D.; and Wang, H.
\newblock 2015.
\newblock Multi-task learning for multiple language translation.
\newblock {\em ACL}.

\bibitem[\protect\citeauthoryear{Galley \bgroup et al\mbox.\egroup
  }{2015}]{galley2015}
Galley, M.; Brockett, C.; Sordoni, A.; Ji, Y.; Auli, M.; Quirk, C.; Mitchell,
  M.; Gao, J.; and Dolan, B.
\newblock 2015.
\newblock delta{BLEU}: A discriminative metric for generation tasks with
  intrinsically diverse targets.
\newblock {\em ACL-IJCNLP}.

\bibitem[\protect\citeauthoryear{Graham, Baldwin, and
  Mathur}{2015}]{graham-baldwin-mathur:2015:NAACL-HLT}
Graham, Y.; Baldwin, T.; and Mathur, N.
\newblock 2015.
\newblock Accurate evaluation of segment-level machine translation metrics.
\newblock {\em NAACL}.

\bibitem[\protect\citeauthoryear{He \bgroup et al\mbox.\egroup
  }{2017}]{HeHeP17-1162}
He, H.; Balakrishnan, A.; Eric, M.; and Liang, P.
\newblock 2017.
\newblock Learning symmetric collaborative dialogue agents with dynamic
  knowledge graph embeddings.
\newblock {\em ACL}.

\bibitem[\protect\citeauthoryear{Hoang, Cohn, and
  Haffari}{2016}]{hoang2016incorporating}
Hoang, C. D.~V.; Cohn, T.; and Haffari, G.
\newblock 2016.
\newblock Incorporating side information into recurrent neural network language
  models.
\newblock {\em NAACL-HLT}.

\bibitem[\protect\citeauthoryear{Hochreiter and
  Schmidhuber}{1997}]{hochreiter1997long}
Hochreiter, S., and Schmidhuber, J.
\newblock 1997.
\newblock Long short-term memory.
\newblock {\em Neural computation} 9(8):1735--1780.

\bibitem[\protect\citeauthoryear{Huang \bgroup et al\mbox.\egroup
  }{2016}]{huang2016attention}
Huang, P.-Y.; Liu, F.; Shiang, S.-R.; Oh, J.; and Dyer, C.
\newblock 2016.
\newblock Attention-based multimodal neural machine translation.
\newblock {\em WMT}.

\bibitem[\protect\citeauthoryear{Li \bgroup et al\mbox.\egroup
  }{2016a}]{li2016diversity}
Li, J.; Galley, M.; Brockett, C.; Gao, J.; and Dolan, B.
\newblock 2016a.
\newblock A diversity-promoting objective function for neural conversation
  models.
\newblock {\em NAACL-HLT}.

\bibitem[\protect\citeauthoryear{Li \bgroup et al\mbox.\egroup
  }{2016b}]{li2016persona}
Li, J.; Galley, M.; Brockett, C.; Gao, J.; and Dolan, B.
\newblock 2016b.
\newblock A persona-based neural conversation model.
\newblock {\em ACL}.

\bibitem[\protect\citeauthoryear{Liu and Perez}{2017}]{DBLP:conf/eacl/LiuP17a}
Liu, F., and Perez, J.
\newblock 2017.
\newblock Dialog state tracking, a machine reading approach using memory
  network.
\newblock {\em EACL}.

\bibitem[\protect\citeauthoryear{Liu \bgroup et al\mbox.\egroup
  }{2015}]{liu-EtAl:2015}
Liu, X.; Gao, J.; He, X.; Deng, L.; Duh, K.; and Wang, Y.-Y.
\newblock 2015.
\newblock Representation learning using multi-task deep neural networks for
  semantic classification and information retrieval.
\newblock {\em NAACL-HLT}.

\bibitem[\protect\citeauthoryear{Liu \bgroup et al\mbox.\egroup
  }{2016}]{notbleu:2016}
Liu, C.-W.; Lowe, R.; Serban, I.; Noseworthy, M.; Charlin, L.; and Pineau, J.
\newblock 2016.
\newblock How {NOT} to evaluate your dialogue system: An empirical study of
  unsupervised evaluation metrics for dialogue response generation.
\newblock {\em EMNLP}.

\bibitem[\protect\citeauthoryear{Luong \bgroup et al\mbox.\egroup
  }{2016}]{luong2015multi}
Luong, M.-T.; Le, Q.~V.; Sutskever, I.; Vinyals, O.; and Kaiser, L.
\newblock 2016.
\newblock Multi-task sequence to sequence learning.
\newblock {\em ICLR}.

\bibitem[\protect\citeauthoryear{Och}{2003}]{mert}
Och, F.~J.
\newblock 2003.
\newblock Minimum error rate training in statistical machine translation.
\newblock {\em ACL}.

\bibitem[\protect\citeauthoryear{Oh and Rudnicky}{2000}]{OhRudnicky2000}
Oh, A.~H., and Rudnicky, A.~I.
\newblock 2000.
\newblock Stochastic language generation for spoken dialogue systems.
\newblock {\em ANLP/NAACL Workshop on Conversational systems}.

\bibitem[\protect\citeauthoryear{Papineni \bgroup et al\mbox.\egroup
  }{2002}]{Papineni2002BLEU}
Papineni, K.; Roukos, S.; Ward, T.; and Zhu, W.-J.
\newblock 2002.
\newblock {\sc Bleu}: a method for automatic evaluation of machine translation.
\newblock {\em ACL}.

\bibitem[\protect\citeauthoryear{Przybocki, Peterson, and
  Bronsart}{2008}]{MetricsMATR:2008}
Przybocki, M.; Peterson, K.; and Bronsart, S.
\newblock 2008.
\newblock Official results of the {NIST} 2008 metrics for machine translation
  challenge.
\newblock In {\em MetricsMATR08 workshop}.

\bibitem[\protect\citeauthoryear{Ratnaparkhi}{2002}]{Ratnaparkhi2002}
Ratnaparkhi, A.
\newblock 2002.
\newblock Trainable approaches to surface natural language generation and their
  application to conversational dialog systems.
\newblock {\em Computer Speech \& Language} 16(3):435--455.

\bibitem[\protect\citeauthoryear{Ritter, Cherry, and
  Dolan}{2011}]{ritter2011data}
Ritter, A.; Cherry, C.; and Dolan, W.~B.
\newblock 2011.
\newblock Data-driven response generation in social media.
\newblock {\em EMNLP}.

\bibitem[\protect\citeauthoryear{Serban \bgroup et al\mbox.\egroup
  }{2015}]{corpora:2015}
Serban, I.~V.; Lowe, R.; Charlin, L.; and Pineau, J.
\newblock 2015.
\newblock A survey of available corpora for building data-driven dialogue
  systems.
\newblock {\em CoRR} abs/1512.05742.

\bibitem[\protect\citeauthoryear{Serban \bgroup et al\mbox.\egroup
  }{2016}]{serban2015hierarchical}
Serban, I.~V.; Sordoni, A.; Bengio, Y.; Courville, A.; and Pineau, J.
\newblock 2016.
\newblock Building end-to-end dialogue systems using generative hierarchical
  neural network models.
\newblock {\em AAAI}.

\bibitem[\protect\citeauthoryear{Shang, Lu, and Li}{2015}]{shang2015neural}
Shang, L.; Lu, Z.; and Li, H.
\newblock 2015.
\newblock Neural responding machine for short-text conversation.
\newblock {\em ACL-IJCNLP}.

\bibitem[\protect\citeauthoryear{Sordoni \bgroup et al\mbox.\egroup
  }{2015}]{sordoni2015}
Sordoni, A.; Galley, M.; Auli, M.; Brockett, C.; Ji, Y.; Mitchell, M.; Nie,
  J.-Y.; Gao, J.; and Dolan, B.
\newblock 2015.
\newblock A neural network approach to context-sensitive generation of
  conversational responses.
\newblock {\em NAACL-HLT}.

\bibitem[\protect\citeauthoryear{Sukhbaatar \bgroup et al\mbox.\egroup
  }{2015}]{sukhbaatar2015}
Sukhbaatar, S.; Weston, J.; Fergus, R.; et~al.
\newblock 2015.
\newblock End-to-end memory networks.
\newblock {\em NIPS}.

\bibitem[\protect\citeauthoryear{Sutskever, Vinyals, and
  Le}{2014}]{sutskever2014sequence}
Sutskever, I.; Vinyals, O.; and Le, Q.~V.
\newblock 2014.
\newblock Sequence to sequence learning with neural networks.
\newblock {\em NIPS}.

\bibitem[\protect\citeauthoryear{Vinyals and Le}{2015}]{vinyals2015neural}
Vinyals, O., and Le, Q.
\newblock 2015.
\newblock A neural conversational model.
\newblock {\em ICML Deep Learning Workshop}.

\bibitem[\protect\citeauthoryear{Wen \bgroup et al\mbox.\egroup
  }{2015}]{wen-EtAl2015}
Wen, T.-H.; Gasic, M.; Mrk\v{s}i\'{c}, N.; Su, P.-H.; Vandyke, D.; and Young,
  S.
\newblock 2015.
\newblock Semantically conditioned {LSTM}-based natural language generation for
  spoken dialogue systems.
\newblock {\em EMNLP}.

\bibitem[\protect\citeauthoryear{Wen \bgroup et al\mbox.\egroup
  }{2016}]{DBLP:journals/corr/WenGMRSVY16}
Wen, T.-H.; Ga\v{s}i\'{c}, M.; Mrk\v{s}i\'{c}, N.; Rojas-Barahona, L.~M.; Su,
  P.-H.; Vandyke, D.; and Young, S.
\newblock 2016.
\newblock Multi-domain neural network language generation for spoken dialogue
  systems.
\newblock {\em NAACL-HLT}.

\bibitem[\protect\citeauthoryear{Wen \bgroup et al\mbox.\egroup
  }{2017}]{Wen2017Latent}
Wen, T.-H.; Miao, Y.; Blunsom, P.; and Young, S.
\newblock 2017.
\newblock Latent intent dialog models.
\newblock {\em ICML}.

\bibitem[\protect\citeauthoryear{Weston \bgroup et al\mbox.\egroup
  }{2016}]{weston2015}
Weston, J.; Bordes, A.; Chopra, S.; Rush, A.~M.; van Merri{\"e}nboer, B.;
  Joulin, A.; and Mikolov, T.
\newblock 2016.
\newblock Towards {AI}-complete question answering: A set of prerequisite toy
  tasks.
\newblock {\em ICLR}.

\bibitem[\protect\citeauthoryear{Weston, Chopra, and Bordes}{2015}]{weston2014}
Weston, J.; Chopra, S.; and Bordes, A.
\newblock 2015.
\newblock Memory networks.
\newblock {\em ICLR}.

\bibitem[\protect\citeauthoryear{Zhao \bgroup et al\mbox.\egroup
  }{2017}]{Zhao2017ACL}
Zhao, T.; Lu, A.; Lee, K.; and Eskenazi, M.
\newblock 2017.
\newblock Generative encoder-decoder models for task-oriented spoken dialog
  systems with chatting capability.
\newblock {\em ACL}.

\end{thebibliography}
\bibliographystyle{aaai}
\end{document}